\documentclass{article}
\usepackage[utf8]{inputenc}

\usepackage{overpic}
\usepackage{amssymb}
\usepackage{amsmath}
\usepackage{graphicx}
\usepackage{booktabs} 
\usepackage{multirow} 
\usepackage{diagbox}
\usepackage[T1]{fontenc} 
\usepackage[style=italian]{csquotes} 
\usepackage{tikz}

\usepackage{color}
\definecolor{turquoise}{cmyk}{0.65,0,0.1,0.3}
\definecolor{purple}{rgb}{0.65,0,0.65}
\definecolor{dark_green}{rgb}{0, 0.5, 0}
\definecolor{orange}{rgb}{0.8, 0.6, 0.2}
\definecolor{red}{rgb}{0.8, 0.2, 0.2}
\definecolor{blueish}{rgb}{0.0, 0.7, 1}
\definecolor{light_gray}{rgb}{0.7, 0.7, .7}
\definecolor{pink}{rgb}{1, 0, 1}
\definecolor{dark_red}{rgb}{0.5, 0, 0}

\newcommand{\hide}[1]{{}} 

\newtheorem{theorem}{Theorem}

\newtheorem{example}[theorem]{Example}
\newtheorem{remark}[theorem]{Remark}

\allowdisplaybreaks

\usepackage{currfile}

\usepackage{blindtext}

\usepackage{parskip}
\renewcommand{\paragraph}[1]{\vspace{2\parskip}\textbf{#1}}

\DeclareMathAlphabet\mathbfcal{OMS}{cmsy}{b}{n}

\newcommand{\CIRCLE}[1]{\raisebox{.5pt}{\footnotesize \textcircled{\raisebox{-.6pt}{#1}}}}

\newcommand{\RR}{\mathbb{R}}
\newcommand{\ZZ}{\mathbb{R}}

\newcommand{\BP}{\mathbb{P}}

\newcommand{\CalU}{\mathcal{U}}
\newcommand{\CalV}{\mathcal{V}}

\newcommand{\CalW}{\mathcal{W}}

\newcommand{\compose}{\circ}
\newcommand{\Union}{\cup}

\newcommand{\Char}{\chi}

\newcommand{\st}{:}

\newcommand{\convolution}{\circledast}

\newcommand{\source}{\mathbb{U}}
\newcommand{\target}{\mathbb{T}}
\newcommand{\set}[1]{\left\{#1\right\}}

\title{Voronoi Convolutional Neural Networks}
\author{Soroosh Yazdani and Andrea Tagliasacchi}
\date{}

\begin{document}
\maketitle
\begin{abstract}
In this technical report, we investigate extending convolutional neural networks to the setting where functions are not sampled in a grid pattern.
We show that by treating the samples as the average of a function within a cell, we can find a natural equivalent of most layers used in CNN.
We also present   an algorithm for running inference for these models exactly using standard convex geometry algorithms.
\end{abstract}
\section{Introduction}
When dealing with real world problems, we usually model them by real functions with an $n$-dimensional \textit{continuous} domain, for instance $f:\RR^2 \rightarrow \RR^3$ for RGB images. For this paper we use $\source$ to denote the domain 
of the function and $\target$ to be the target space.
However, continuous domains are not directly usable in computation.
As a result, we usually approximate $f:\source \rightarrow \target$ with
an approximate function $f_\CalU : \CalU \rightarrow \target$ where $\CalU$ is
a finite set and $f_\CalU$ is a discretization of $f$.
\begin{example}
\label{std_quant}
  When $\source = [0,1]^2$ and $\target = \RR$, it is standard to set 
  \[ \CalU = \set{ (i/w, j/h) \in \RR^2 \st i,j\in\ZZ, \, 0\leq i <w, \, 0\leq j < h}. \]
  We can set $f_\CalU$ by a restriction map. In this instance, any function
  $f:\source \rightarrow \target$ can be approximated as a two dimensional array,
  which is an standard representation of gray scale images, for instance.
\end{example}
If $f$ is compactly supported, then we can assume that $\source \!\subset\! [0, 1]^s$ and  $\target\!\subset\!\RR^t$. Using the standard grid discretization
above, the function can be described in a $(s\!+\!1)$-dimensional tensor.
However, one could ask: \enquote{could we replace grid discretizations with something more general?}

In this paper, we propose using piecewise constant approximation on a Voronoi~\cite[Ch.~7]{book}
partition of the domain for this purpose.
We chose this approximation since it is the building block for a novel generalization of the standard convolutional neural network, which we call {\em Voronoi Convolutional Neural Networks}. 

\section{Partitioning}
Let $f:\source \rightarrow \RR^n$, and let $S \subset \source$. In this
paper we use the notation
$$f(S) = \frac{1}{|S|} \int_S f(x) \: dx, $$
where $|S| = \int_S dx$ is the volume of $S$.

Let $\CalU = \{S_k\}$ be a \textit{partitioning} of $\source$, the domain of $f$, such that $\CalU = \Union \: S_k$ and $S_i \cap S_j = \emptyset$ for all $i \neq j$.
Then define the piecewise constant approximation of $f$ on $\CalU$, denoted
by $f_\CalU$, by
$$
f_\CalU(x) = \sum_{S \in \CalU} f(S) \Char_{S}(x)
$$
where $\Char_{S}(x)$ is the indicator function for
the set $S$ ($\Char_{S}(x)=1$ if $x \in S$, otherwise, it is $0$).
Note that $f_\CalU$ is constant on each set $S \in \CalU$. As such, 
for any $S \in \CalU$ and $s \in S$ we have $f_\CalU(s)=f_\CalU(S)$. 

We would like to formulate the typical neural network operations on the space of functions of the form~$f_\CalU: \CalU \rightarrow \RR^n$.

\begin{example}
  \label{avg_quant}
  Let $\source = [0,1)^2$. The standard $w \times h$ grid partitioning of
  $\source$ is given by $\CalU = \set{ S_{i,j} \st 0<i<w, 0<j<h}$ where
  \[ S_{i,j}=\set{(x,y) \in \source \st \frac{i}{w}\leq x< \frac{i+1}{w}, \frac{j}{h} \leq y < \frac{j+1}{h} }. \]
  Then for any $f:[0,1]^2 \rightarrow \RR^n$, we can describe $f_\CalU$
  as an $(w, h, n)$-tensor.
\end{example}
Note that the approximations of $f$ in examples \ref{std_quant} and \ref{avg_quant} are defined by $(w, h, n)$-tensors. However, in case
of example \ref{std_quant}, the entries of the tensor are the value of the
underlying function, while in example \ref{avg_quant}, the entries are the
average value of the function over the cell $S_{i, j}$.
In general, if $\CalU$ has $k$ elements in it, any function
$f_\CalU : \CalU \rightarrow \RR^n$ can be described as a $(k,n)$-tensor.

Amongst possible partitioning schemes, we are particularly interested in \textit{Voronoi} partitioning, as \CIRCLE{1} it is a strict superset of grid partitioning, and \CIRCLE{2} it simplifies the discretization of convolution by decomposing a domain into \textit{convex} polytopes.

\begin{remark}
 We chose piecewise constant approximation, since it makes the math work 
 later in this paper.
 However, other class of functions can work here as well. For instance,
 if the partitions are all polyhedra, one can use generalized barycentric
 coordinates to get a continuous approximation of the underlying function.
\end{remark}


\section{Convolutions}
Let $\source = \RR^a$, and
consider two functions $f: \source \rightarrow \RR^m$, and $\kappa: \source \rightarrow \RR^{m \times n}$ (i.e. $\kappa(x)$ is an $m \times n$ matrix).
Recall the definition of convolution $f$ and $\kappa$, denoted by
$\kappa \convolution f : \source \rightarrow \RR^n$:
\begin{equation} \label{eq:conv}
(\kappa \convolution f)[x] = \int_\source \kappa(x-\tau) \cdot f(\tau) \: dt.
\end{equation}
Consider three partitions $\mathcal{U}, \mathcal{V},$ and $\mathcal{W}$ of $\source$.
Let us now assume that $f=f_\mathcal{U}$ (i.e.
$f$ is constant on each $U \in \mathcal{U}$) and $\kappa = \kappa_\mathcal{V}$.
We wish to compute the approximation of $(\kappa \convolution f)$ on $\mathcal{W}.$ In particular, for $W \in \mathcal{W}$:
\begin{eqnarray*}
(\kappa \convolution f)(W) 
&=& \frac{1}{|W|} \int_{W} (\kappa \convolution f)[w] \: dw \\
&=& \frac{1}{|W|} \int_{W} \int_\source \kappa(w-u) \cdot f(u) \: dw du  \\
&=& \frac{1}{|W|} \int_{W} \sum_{U \in \CalU} \int_U \kappa(w-u) \cdot f(U) \:  dw du \\
&=& \frac{1}{|W|} \sum_{U \in \CalU} \left(\int_W \int_U \kappa(w-u) \: dw du \right) \cdot f(U) \\
&=& \frac{1}{|W|} \sum_{U \in \CalU} K_{U,W} \cdot f(U)
\end{eqnarray*}
where in the expression above:
\begin{eqnarray*}
K_{U,W} 
&=& \int_W \int_U \kappa(w-u) \: du dw \\
&=& \int_{\RR^a} \int_{\RR^a} \kappa(w-u) \Char_U(u) \Char_W(w) \: du dw.
\end{eqnarray*}
Recall, as before, $\Char_S$ is the indicator function for set $S$. Also, note
that $K_{U,W}$ is an $m \times n$ matrix and $f(U)$ is an $m$-dimensional
vector.

%

\section{Volume computation}
We can find numerical approximation for $K_{U,W}$ in general case.
However, when $U, V,$ and $W$ are all {\em convex} polytopes, then we
can compute $K_{U,W}$ using convex geometry algorithms.
First note that we can rewrite $K_{U,W}$ (for all sets $U$ and $V$) as
\begin{align*}
K_{U,W} 
&= \int_{\RR^a} \int_{\RR^a} \kappa(w-u) \Char_U(u) \Char_W(w) \: du dw \\
&= \int_{\RR^a} \int_{\RR^a} \kappa(x) \Char_U(u) \Char_W(x+u) \: du dx \\
&= \int_{\RR^a} \kappa(x) \left( \int_{\RR^a} \Char_U(u) \Char_{(W-x)}(u) \: du \right) dx \\
&= \int_{\RR^a} \kappa(x) \left( \int_{\RR^a} \Char_{U \cap (W-x)}(u)  \: du \right) dx \\
&= \int_{\RR^a} \kappa(x) \: | U \cap (W-x) | \: dx.
\end{align*}
Where $|\cdot|$ indicates the volume of a partition.
In particular note that
\begin{eqnarray*}
K_{U,W} & = & \int_{\RR^a} \kappa(x) \: |U \cap (W-x)| \: dx \\
& = & \sum_V \int_{V} \kappa(V) \: |U \cap (W-x)| \: dx \\
&=& \sum_V \kappa(V) \int_{\RR^a} \int_{\RR^a} \Char_V(x) \Char_U(y) \Char_W(x+y) dx dy \\
&=& \sum_V \kappa(V) K_{U,W}^V
\end{eqnarray*}
%
%
Now assume $U$, $V$ and $W$ to be \textit{convex} polytopes.
Then we can define a convex polytope in $\RR^{2a}$ so that $K_{U,W}^V$ is the volume of the given polytope. 
In particular, assume that each set is defined as the intersection of halfspaces:
\begin{align}
U = \bigcap_i H_i, \quad
V = \bigcap_j H'_j, \quad
W = \bigcap_k H''_k.
\end{align}
Then we can rewrite
\begin{eqnarray*}
K_{U,W}^V
&=& \int_{\RR^a} \int_{\RR^a} \Char_V(x) \Char_U(y) \Char_W(x+y) dx dy \\
&=& \int_{\RR^a} \int_{\RR^a} 
\left(\prod_i \Char_{H_i}(x)\right) 
\left(\prod_j \Char_{H'_j}(y)\right) 
\left(\prod_k \Char_{H''_k}(x+y)\right) dx dy.
\end{eqnarray*}
Note that each term in the product is the composition of a Heaviside function
with a linear function, hence the whole product is the indicator function
of a convex polytope in $\RR^{2a},$ and as such we can compute its volume
using standard convex polytope algorithms.

\section{Voronoi Convolutional Network}
Let $\BP=\{p_n \in \source\}$ be a set of points.
Then the Voronoi cell $U_p$ is defined as
\[ U_p = \{ x \in \RR^a \st d(x,p) \leq d(x,q) \, \forall q \in \BP\}.\]
Note that the Voronoi cells $U_p$ form a partitioning:
$$
\CalU_{\BP} = \{ U_p \st p \in \BP\}.
$$
Furthermore, it is well known that each Voronoi cell is a \textit{convex} polytope, and there are libraries that given a set of points, will compute the Voronoi cells as a convex polytope. 
We used \verb|scipy.spatial| library for our computation.

We can now define our {\em Voronoi Convolutional Networks}.
Each layer of such a network is given by a Voronoi partitioning
of $\source$, denoted by $\CalU$, and a function $f:\CalU \rightarrow \RR^m$.
Note that a layer $(\CalU, f)$ are encoded by set of $k$ points in $\source$ and a
tensor of shape $k \times m$ for the function $f$.
We define few transition layers, similar to the ones used in CNNs.
\begin{itemize}
    \item {\bf Convolutions:} Given a set of convex spaces
    $\CalV=\{V_1, \ldots V_e\}$, we define $\CalV$-convolution as a
    matrix valued function
    $\kappa: \CalV \rightarrow \RR^{m \times n}$.
    Given a layer in VCNN $(\CalU, f)$, a convolution
    $\kappa$, and a partitioning $\CalW$, we get a new layer
    $(\CalW, f \convolution \kappa)$ by
    \begin{align}
    (f \convolution \kappa)(W) = \sum_{U,V} K_{U,W}^V \kappa(V) \cdot f(U).
    \label{ConvFormula}
    \end{align}
    Given an activation function $\sigma :\RR \rightarrow \RR$ as well,
    we get the new layer
    $(\CalW, \sigma \compose (f \convolution \kappa) )$.
    \item {\bf (Average) Pooling:} Given a layer in VCNN $(\CalU, f)$ and a partitioning
    $\CalW$, then the pooling of $f$ to $\CalW$ is $(\CalW, g)$ where
    \[g(W) = \sum_V |V \cap W|f(V). \]
    \item {\bf Mixup / $(1 \times 1)$-convolution:} The equivalent to the $1 \times 1$
    convolution in $2D$-CNNs, is just applying a matrix $M$ to $f$. In 
    particular, for layer $(\CalV, f)$, applying the MixUp we get
    $(\CalV, M \compose f)$.
    
    \item {\bf Concat}: We can concatenate two layers if they are both defined
    on the same domain. That is given $(\CalV, f)$ and $(\CalV, g)$, we define
    the concatenation of these two layers to be $(\CalV, f \bigoplus g)$.
\end{itemize}
We define a VCNN as a network that is built as a stack of the layers above.
Notice that a VCNN is differentiable if the transition layers are
differentiable. As such, we can apply back propagation to train a network
for solving particular problems.

The following two examples shows that VCNNs are generalization of standard
CNNs:

\begin{example}[1D-Conv]
  Assume $\CalU=\CalW = \{ [i,i+1) \st i \in \ZZ\}$, let
  $f:\CalU \rightarrow \RR^n$, and consider input layer $(\CalU, f)$.
  Let $\CalV = \{[0,1), [1,2), [2,3)\}$, and $\kappa : \CalV \rightarrow \RR^{n \times m}$ be the $\CalV$ convolution.
  For simplicity we denote $f([i,i+1))$ by $f_i$ and denote
  $\kappa([i,i+1)]$ by $M_i$ (so $\CalV$ convolution is defined by
  $M_0, M_1$, and $M_2$).
  Then applying $\kappa$ to $(f_i)$ we get
  \[ g_i = (M_0 f_i + f_{i-1}(M_0 + M_1) +f_{i-2}(M_1 + M_2) + M_2 f_{i-3})/2. \]
  Notice that this is slightly different than the 1D convs with kernel
  size of $3$ with $m$ filters.
\end{example}

\begin{remark}
 Note that the pooling layers in VCNN can go from any partitioning to
 any other partitioning. In particular, we can go from more sparse
 partitioning to denser partitioning.
\end{remark}
\section{Conclusion}
In this note we showed how one can extend CNNs to the case where the sampling
of is not done in a grid pattern. We developed our network by considering
convolutions on continuous functions, and discretizing the functions on
Voronoi cells. The networks we have presented here can be implemented in standard
deep learning frameworks such as TensorFlow or PyTorch, under the assumption
that the Voronoi cells used in each layers do not change.
We note that when the cell $V$ is fairly small compared to cells
$\CalU$ and $\CalW$, then for many choices of $U$ and $W$ we expect
$K_{U,W}^V=0$, and for cases were $\CalU$ and $\CalW$ have large number
of cells, we need to implement an algorithm to identify those quickly
for efficiency reasons. We did not investigate such algorithms in this paper.

Note that to really get the benefit of such networks, we really like to allow
the points defining the Voronoi cells in each layer to move freely. To achieve
that, we need to be able to compute the derivative of volume computation
$K_{U,W}^V$ with respect to the Voronoi cells, which we did not undertake in
this paper.
On the other hand, one can treat $K_{U,W}^V$ as trained variables. In that
case, the resulting network can be interpreted as a multihead attention network,
where the attention of Query and Key should be interpreted as $K_{U,W}^V$.
It would be interesting to see if the attention matrices in standard
attention networks carry a geometric interpretation similar to the volume
of Voronoi cells.
{
    \small
    \bibliographystyle{ieee_fullname}
    \bibliography{main}
}
\end{document}